\documentclass{article}

\usepackage{arxiv}

\usepackage[utf8]{inputenc} % allow utf-8 input
\usepackage[T1]{fontenc}    % use 8-bit T1 fonts
\usepackage{hyperref}       % hyperlinks
\usepackage{url}            % simple URL typesetting
\usepackage{booktabs}       % professional-quality tables
\usepackage{amsfonts}       % blackboard math symbols
\usepackage{nicefrac}       % compact symbols for 1/2, etc.
\usepackage{microtype}      % microtypography
\usepackage{graphicx}
\usepackage[numbers,sort&compress]{natbib}
\usepackage{doi}

\usepackage{graphicx}%
\usepackage{multirow}%
\usepackage{amsmath,amssymb,amsfonts}%
\usepackage{amsthm}%
\usepackage{mathrsfs}%
\usepackage[title]{appendix}%
\usepackage{xcolor}%
\usepackage{textcomp}%
\usepackage{manyfoot}%
\usepackage{booktabs}%
\usepackage{bm} % for bold math symbols
\usepackage{algorithm}%
\usepackage{algorithmicx}%
\usepackage{algpseudocode}%
\usepackage{listings}%
\usepackage{makecell}

\title{Restoration Adaptation for Semantic Segmentation on Low Quality Images}

\date{} 					% Or removing it

\author{
  \begin{tabular}{c}
    Kai Guan$^{1,2}$, Rongyuan Wu$^1$, Shuai Li$^1$, Wentao Zhu$^2$, 
    Wenjun Zeng$^{2,*}$, Lei Zhang$^{1,}$\thanks{Corresponding authors.} \\[1ex]
    \small \texttt{$^1$The Hong Kong Polytechnic University \quad $^2$Eastern Institute of Technology, Ningbo} \\
    \small \texttt{\{kai11.guan, rong-yuan.wu, novak.li\}@connect.polyu.hk} \\
    \small \texttt{\{wtzhu, wzeng-vp\}@eitech.edu.cn, cslzhang@comp.polyu.edu.hk}
  \end{tabular}
}

\renewcommand{\shorttitle}{\textit{arXiv} Template}

\hypersetup{
pdftitle={A template for the arxiv style},
pdfsubject={q-bio.NC, q-bio.QM},
pdfauthor={David S.~Hippocampus, Elias D.~Striatum},
pdfkeywords={First keyword, Second keyword, More},
}

\begin{document}
\maketitle

\begin{abstract}
In real-world scenarios, the performance of semantic segmentation often deteriorates when processing low-quality (LQ) images, which may lack clear semantic structures and high-frequency details. Although image restoration techniques offer a promising direction for enhancing degraded visual content, conventional real-world image restoration (Real-IR) models primarily focus on pixel-level fidelity and often fail to recover task-relevant semantic cues, limiting their effectiveness when directly applied to downstream vision tasks. Conversely, existing segmentation models trained on high-quality data lack robustness under real-world degradations. In this paper, we propose Restoration Adaptation for Semantic Segmentation (RASS), which effectively integrates semantic image restoration into the segmentation process, enabling high-quality semantic segmentation on the LQ images directly. Specifically, we first propose a Semantic-Constrained Restoration (SCR) model, which injects segmentation priors into the restoration model by aligning its cross-attention maps with segmentation masks, encouraging semantically faithful image reconstruction. Then, RASS transfers semantic restoration knowledge into segmentation through LoRA-based module merging and task-specific fine-tuning, thereby enhancing the model's robustness to LQ images. To validate the effectiveness of our framework, we construct a real-world LQ image segmentation dataset with high-quality annotations, and conduct extensive experiments on both synthetic and real-world LQ benchmarks. The results show that SCR and RASS significantly outperform state-of-the-art methods in segmentation and restoration tasks. Code, models, and datasets will be available at \url{https://github.com/Ka1Guan/RASS.git}.
\end{abstract}

% keywords can be removed
\section{Introduction}\label{intro}
% \end{links}

Semantic segmentation \cite{cheng2021per, cheng2022masked, guo2018review, strudel2021segmenter} is one of the prominent research topics in computer vision, while pixel-level categorization of visual content is crucial for various applications, such as scene understanding \cite{mo2022review, hofmarcher2019visual, chen2023fpr}. Starting with Fully Convolutional Networks (FCNs) \cite{long2015fully}, pixel-by-pixel classification models have been widely used in semantic segmentation. CNN-based models achieve segmentation by extracting hierarchical features from images through convolutional layers, followed by downsampling and upsampling processes that enable dense pixel-wise classification \cite{chen2018encoder, ronneberger2015u, badrinarayanan2017segnet}. Later, transformer-based models, such as SegFormer \cite{xie2021segformer} and Mask2Former \cite{cheng2022masked}, have achieved state-of-the-art (SOTA) performance on semantic segmentation tasks. Dividing images into patches (tokens) and using self-attention to capture global context and relationships between tokens, these architectures allow for a more nuanced categorization of image regions \cite{cheng2022masked, ni2024context, cheng2021per}.

\begin{figure}[t]
  \centering
  \includegraphics[width=0.9\columnwidth]{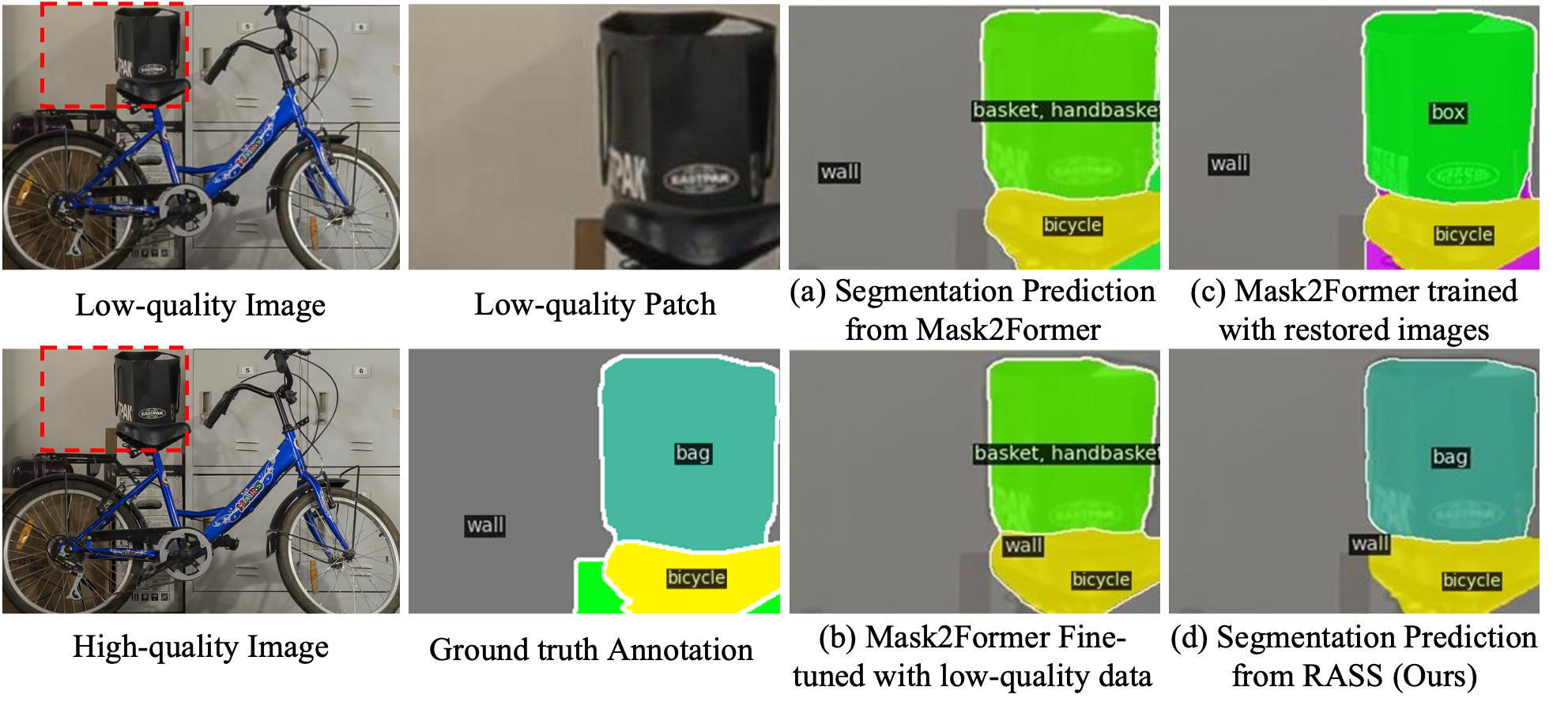}
  \caption{Our RASS framework integrates semantic-guided recovery into the segmentation backbone for robust parsing of low-quality images. (a) Mask2Former~\cite{cheng2022masked} trained on high-quality data fails to segment degraded regions (\textit{e.g.}, blurred objects like bag). (b) Fine-tuning on low-quality data recovers coarse structure but struggles with ambiguous targets due to inconsistent features. (c) Restoration as preprocessing can enhance image clarity, but generation without semantic guidance may lead to misclassification (\textit{e.g.}, box). (d) RASS adaptively incorporates restoration into segmentation, better capturing degraded objects while avoiding error propagation from disjoint pipelines.}
  \label{fig1}
  %\vspace{-0.3cm}
\end{figure}

Current segmentation models, including those CNN-based \cite{long2015fully, chen2017deeplab} and Transformer-based \cite{xie2021segformer, cheng2022masked} ones, are primarily trained on clear standard-quality (SQ) images. However, the captured images in real-world applications are often subjected to various image degradations \cite{liang2021swinir, zhang2023ingredient}. Although \cite{chen2024robustsam} extended the general segmenter SAM \cite{kirillov2023segment} to multiple single degradation type images, including blur, noise, \textit{etc}, real-world degradation types are often complex and unpredictable, and research on semantic segmentation under such conditions remains limited. Some feature-refinement methods \cite{couairon2025jafar, fu2024featup} have been proposed to improve segmentation by increasing spatial resolution, but their effectiveness depends on reliable spatial cues that may be absent under complex corruptions. As illustrated in Fig. \ref{fig1} (a), directly applying current segmentation models such as Mask2Former~\cite{cheng2022masked} to LQ images will struggle in extracting robust features for accurate segmentation. One intuitive solution is to re-train these models on LQ images. However, on the one hand, the architecture design of existing segmentation networks is not suitable for LQ image feature extraction; on the other hand, this solution ignores the exploitation of high-quality (HQ) image priors, which can be used to improve the quality of LQ images and consequently enhance the segmentation performance. As illustrated in Fig. \ref{fig1} (b), the fine-tuned Mask2Former on LQ images still yields unsatisfactory results.

Restoring HQ images from LQ images has been a long-standing research topic in low-level vision \cite{liang2021swinir, zhang2023ingredient, kong2022reflash}, while real-world image restoration (Real-IR) techniques have recently made significant progress \cite{chen2023hat, valanarasu2022transweather, gou2023test, luo2023controlling}. Using image restoration as a pre-processing step for semantic segmentation training is a straightforward idea: first enhance the quality of degraded images and then feed them into the segmentation model as training data. However, with this approach, the segmentation results will largely depend on the reliability of the restoration process, leading to suboptimal segmentation performance \cite{guo2019degraded}. As illustrated in Fig.~\ref{fig1}(c), the `bag’ on the bicycle is mistakenly restored with the shape and texture of a `box’, despite the overall improvement in image clarity. This over-generation of incorrect details, however, undermines the subsequent segmentation, leading to misclassification.

Recently developed Stable Diffusion (SD) \cite{rombach2021highresolution} based generative Real-IR models have demonstrated strong capability to restore HQ images from LQ inputs \cite{wu2024one, wu2024seesr, wang2023exploiting, lin2023diffbir, sun2025pixel}. While Wu \textit{et al.} \cite{wu2024seesr} showed that SD-based generative restoration can benefit LQ image segmentation tasks, the over-generated details can distort the underlying image structure and introduce noticeable visual artifacts. Furthermore, these SD-based Real-IR models often suffer from mismapping and weak spatial alignment due to their reliance on text prompts without explicit semantic or spatial guidance, ultimately resulting in semantically incorrect outputs. Such limitations significantly hinder their effectiveness in downstream segmentation tasks. Some researchers have explored the integration of restoration into the segmentation model training process. For example, Niu \textit{et al.} \cite{niu2020effective} cascaded the restoration network and the segmentation network during training, while Lee \textit{et al.} \cite{lee2024frest} proposed FREST, which aims to address the challenge of adverse conditions, by learning an additional embedding space. However, these approaches typically rely on auxiliary restoration modules or multi-stage training objectives to learn feature alignment, rather than leveraging semantically rich pre-trained generative priors for a unified segmentation model.

In contrast to Real-IR, which aims to reconstruct perceptually realistic HQ images from real-world LQ inputs, our objective is to directly obtain accurate semantic segmentation maps from these LQ images containing composite and unknown degradations, rather than merely addressing images with noise or other single degradation types. In this paper, we propose Restoration Adaptation for Semantic Segmentation (RASS) on LQ images, a novel framework that effectively integrates image enhancement into the segmentation process to handle de-degradation and segmentation simultaneously, achieving robustness to real-world degradations via high-order synthesis generalization. RASS does not treat restoration as a pre-processing step, but instead fuses pretrained restoration parameters into the segmentation model to enhance robustness to LQ conditions. Although training remains sequential, RASS fuses restoration knowledge into the backbone via LoRA so that, at inference, the segmentation head operates on ``cleaned'' latent features, thereby avoiding error propagation caused by re-encoding intermediate restored pixels. To build a unified framework of restoration and segmentation, we adopt the SD model as backbone. While existing SD-based restoration models focus on fine-grained textures, they often damage structure and cause artifacts. We propose a Semantic-Constrained Restoration (SCR) model to mitigates these issues by guiding restoration with semantic awareness. RASS then leverages LoRA-based find-tuning to transfer restoration knowledge into segmentation, enhancing robustness to degradation without retraining from scratch. The main contributions of this work are summarized as follows:
\begin{itemize}
    \item We propose RASS, which, to the best of our knowledge, is the first framework designed to address semantic segmentation of real-world LQ images.
    \item We propose SCR, injecting segmentation priors by aligning cross-attention maps with semantic masks to recover visually and semantically consistent content.  
    \item We present a carefully annotated real-world LQ segmentation dataset for comprehensive evaluation. Extensive experiments show RASS outperforms existing models on both synthetic and real-world data, with SCR also improving results on the Real-IR benchmark.
\end{itemize}

\section{Related Work}
% \vspace{-4pt}
We conduct the literature review from the perspectives of image segmentation and image restoration. Specifically, we supplement the literature review of SQ image segmentation, as it is the foundation and technical support for LQ image segmentation models. Given the strong capabilities of SD-based Real-IR models, our review of image restoration primarily focuses on this line of research.

\textbf{Standard Quality Image Segmentation:}
Segmentation on SQ images has been extensively studied. Starting from FCN \cite{long2015fully}, which enables pixel-wise prediction using fully convolutional networks, many CNN-based methods have been developed to improve accuracy and efficiency \cite{ronneberger2015u, badrinarayanan2017segnet, chen2017deeplab}. Transformer-based models, such as SegFormer \cite{xie2021segformer},  significantly improve complex scene segmentation by capturing long-range interactions and global context. Mask2Former \cite{cheng2022masked} has achieved a breakthrough in decoder design, which integrates Transformer-based attention mechanisms into a unified framework for instance and demonstrate that, setting new SOTA benchmarks across various datasets. More recently, efficiency-oriented models like FeedFormer \cite{shim2023feedformer}, SeaFormer \cite{wan2025seaformer++}, and CGRSeg \cite{ni2024context} further reduce computation via low-level features, axial attention, or context-guided reconstruction. Despite the impressive performance of current segmentation models, they are primarily trained on SQ images and lack robustness to LQ images, which are common in practical applications. Fine-tuning these models with LQ images may improve robustness but cannot fully address the issue of image degradation. This is because the design of SQ segmentation models relies on extracting discriminative features from images, which are obscured by the unknown degradation in LQ images.

\textbf{Low Quality Image Segmentation:}
Recent efforts have explored LQ segmentation. One intuitive approach is to restore images before segmentation. For instance, SR4IR \cite{kim2024beyond} adopts task-guided super-resolution to improve segmentation performance. However, restoration as a pre-processing step is often sub-optimal. Another strategy is using knowledge distillation to extract features from a model trained on HQ images to assist segmentation on degraded images. Guo \textit{et al.} \cite{guo2019degraded} and Endo \textit{et al.} \cite{endo2023semantic} used models trained on clean images as teacher networks to assist LQ image segmentation models. Yet, performance is limited by the teacher’s capacity and domain gap. More recently, some researchers have attempted to collaboratively perform semantic segmentation and image restoration. Lee \textit{et al.} \cite{lee2024frest} proposed FREST for semantic segmentation under adverse conditions, which learns a condition-specific embedding space to restore features in degraded images, simulating normal conditions. Nevertheless, degradation types under adverse environmental conditions are more predictable than in real-world scenarios, while introducing an additional restoration module increases training cost.
%Xia \textit{et al.} \cite{xia2019cooperative} proposed SGAda and EGSyn, which train separate models to refine the segmentation and restoration results, respectively. These models are then jointly trained to adapt to segmentation tasks under challenging environmental conditions. Similarly,

\textbf{Real-World Image Restoration:}
Recent advances have explored diffusion models for real-world image restoration, particularly leveraging pre-trained text-to-image (T2I) models like Stable Diffusion for their strong generative priors \cite{wang2023exploiting, wu2024seesr, wang2024sinsr, wu2025one}. Early methods such as StableSR \cite{wang2023exploiting} and DiffBIR \cite{lin2023diffbir} introduce trainable encoders or two-stage pipelines to adapt SD for LQ inputs. SeeSR \cite{wu2024seesr} enhances semantic alignment using degradation-aware prompts and semantic cues, while OSEDiff \cite{wu2025one} improves efficiency via direct LQ input for one-step inference. PiSA-SR \cite{sun2025pixel} separates pixel- and semantic-level objectives using dual LoRA modules, enabling controllable restoration. However, these models rely heavily on text prompts without explicit semantic or spatial guidance, often resulting in misaligned or semantically incorrect outputs when guidance is insufficient, which may negatively impact downstream tasks like semantic segmentation.

\section{Methodology}

\subsection{Framework Overview}
Based on the above literature review, our motivation lies in the observation that image restoration and segmentation are two inherently interconnected tasks, yet they are often treated independently in existing works. Conventional image restoration methods primarily emphasize pixel-level fidelity, neglecting the recovery of task-relevant semantic structures, which limits their utility in supporting downstream vision tasks. On the other hand, segmentation models are typically trained on high-quality images and lack robustness against real-world degradations, resulting in substantial performance degradation when applied to LQ inputs. %Moreover, existing approaches rarely consider a unified framework that jointly addresses restoration and segmentation, thereby overlooking the potential for mutual enhancement between the two. These limitations motivate us to explore a unified solution that simultaneously restores image quality and preserves semantic understanding, especially under complex real-world degradations.

\begin{figure*}[t]
  \centering
  \includegraphics[width=\textwidth]{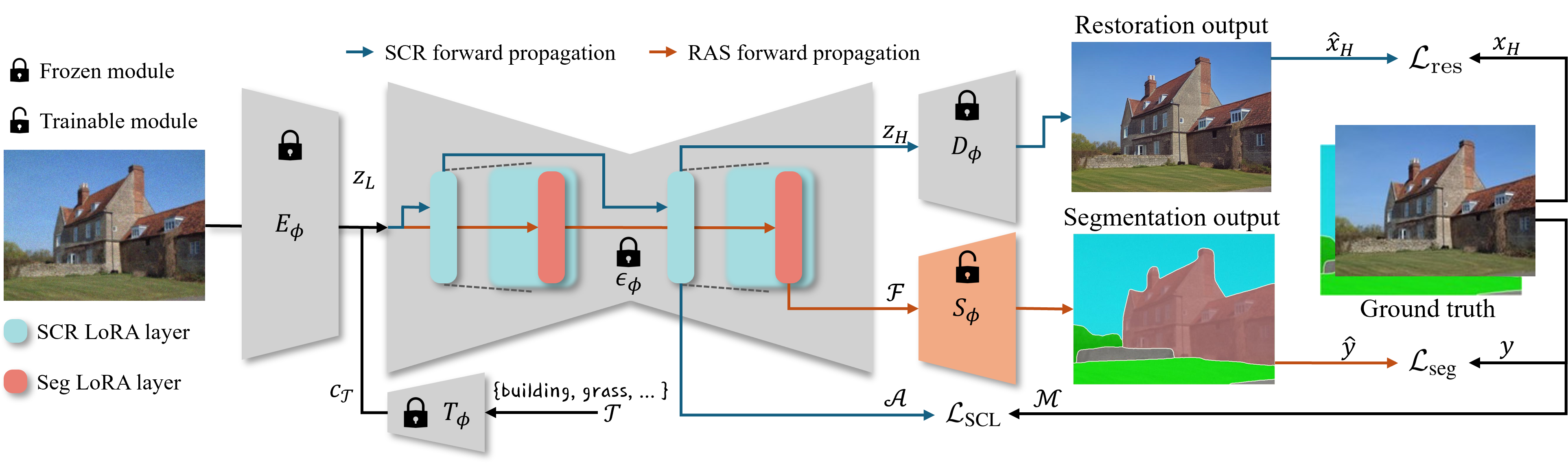}
  \caption{Overview of our RASS training framework. A pre-trained SD model is used as the backbone. In the first stage, the SCR model is trained by injecting trainable SCR LoRA layers into the pretrained diffusion network \( \epsilon_\phi \). The LQ image is passed through the frozen VAE encoder \( E_\phi \), a LoRA finetuned diffusion network \( \epsilon_\phi \), and frozen VAE decoder \( D_\phi \) to generate the restored HQ image \( \hat{\bm{x}}_H \), the given text prompt \(\mathcal{T}\) is processed by the frozen text encoder \( T_\phi \) to obtain the corresponding text embedding \( c_\mathcal{T} \). The restoration is supervised by the loss \(\mathcal{L}_{\mathrm{res}}\) between \( \hat{\bm{x}}_H \) and the ground truth \( \bm{x}_H \), along with a Semantic-Constraint loss \(\mathcal{L}_{\mathrm{SCL}}\) computed from the cross-attention maps \( \mathcal{A} \) in SCR LoRA and the semantic masks \( \mathcal{M} \). In the second stage, the learned SCR LoRA weights are merged and used to initialize new trainable Segmentation (Seg) LoRA layers to train the RAS model. Internal features \( \mathcal{F} \) from \( E_\phi \) and \( \epsilon_\phi \) are fed into a trainable segmentation head \( S_\phi \), with segmentation loss \(\mathcal{L}_{\mathrm{seg}}\) to guide the training process. RASS transfers the restoration knowledge to the segmentation task through LoRA-based module merging and task-specific fine-tuning, thus achieving robust segmentation of LQ images.
}
  \label{fig3}
  % \vspace{-0.3cm}
\end{figure*}

The framework of our proposed Restoration Adaptation for Semantic Segmentation (RASS) method is shown in Figure \ref{fig3}. Our framework includes two core models, Semantic-Constrained Restoration (SCR) model and Restoration Adaptation Segmentation (RAS) model. The two models share a backbone, a T2I model with strong semantic priors. We build our framework with LoRA-based models to complete different tasks. In the first stage, we first train SCR. Considering that T2I models usually have problems with mismapping and poor spatial correspondence due to the lack of detailed spatial guidance, we design a semantic-constrained loss to guide the model to accurately find the target area in response to the text. In the second stage, we design a restoration adaptation strategy that effectively transfers restoration knowledge into the segmentation process by merging the LoRA model, thereby enhancing the robustness of RASS to LQ images without introducing additional modules and parameters. RASS does not iteratively use improved segmentation to refine restoration; instead, its  restoration process is explicitly guided by semantic cues rather than being purely low-level or task-agnostic.

\subsection{Semantic-Constrained Restoration Model}
\label{sec:scr}

%Following the OSEDiff \cite{wu2024one}, a one-step diffusion-based restoration model, as the baseline for our restoration module, 
The recently proposed OSEDiff~\cite{wu2024one} utilizes a single-step diffusion process for end-to-end Real-IR training, enabling the adoption of a residual learning strategy. We denote the LQ and HQ images as \( \bm{x}_L \) and \( \bm{x}_H \), respectively, and their corresponding latent representations as \( \bm{z}_L \) and \( \bm{z}_H \). Let \( E_\phi \), \( \epsilon_\phi \), and \( D_\phi \) represent the VAE encoder, latent diffusion network, and VAE decoder of a pretrained SD model, respectively, where \( \phi \) denotes the model parameters. To adapt SD to our task, we freeze the VAE and introduce trainable restoration LoRA modules (\textit{i.e.}, SCR LoRA in Figure~\ref{fig3}) into the diffusion network. We fine-tune \( \epsilon_\phi \) to \( \epsilon_{\theta_{scr}} \) using LoRA~\cite{hu2022lora}, where \( \theta_{scr} = \{\phi, \Delta \theta_{scr}\} \) and \( \Delta \theta_{scr} \) is the SCR-specific LoRA parameter set. Since SD is a text-conditioned model, we extract the text embedding \( c_\mathcal{T} \) from the input description \( \mathcal{T} \)~\cite{rombach2021highresolution}, perform the noise prediction as \( \hat{\epsilon} = \epsilon_\theta(z_t; t, c_\mathcal{T}) \), in one-step diffusion model (\textit{i.e.}, timestep \(t=1\)), so it can be directly indicated as \( \hat{\epsilon} = \epsilon_{\theta_{scr}}(z_L, c_\mathcal{T}) \). The HQ latent can then be estimated by \(z_{H} = z_L - \hat{\epsilon} \). %and decoded by the VAE decoder with \(x_{H} = D_\phi (z_{H})\).
The general restoration loss function for training SCR LoRA is expressed as follows:
\begin{equation}
\mathcal{L}_{\mathrm{res}}\left( \hat{\bm{x}}_{H},\bm{x}_{H}\right) = 
\lambda_{\mathrm{MSE}}\mathcal{L}_{\mathrm{MSE}} + 
\lambda_{\mathrm{LPIPS}}\mathcal{L}_{\mathrm{LPIPS}} ,
\end{equation}
where $\mathcal{L}_{\mathrm{MSE}}$ is the pixel-wise mean squared error between the restored image $\hat{\bm{x}}_{H}=D_\phi (z_{H})$ and ground truth $\bm{x}_{H}$, encouraging accurate low-level reconstruction. $\mathcal{L}_{\mathrm{LPIPS}}$ denotes the perceptual loss based on deep features from a pretrained network, promoting visually realistic results. We set $\lambda_{\mathrm{MSE}}=1$ and $\lambda_{\mathrm{LPIPS}}=2$ to balance pixel-level and perceptual quality, following \cite{wu2024one}.

To incorporate fine-grained semantic priors into the restoration process, we introduce a Semantic-Constrained Loss (SCL) that guides the SCR LoRA training by aligning cross-attention responses with object-level semantic information. While the base SCR model operates within the latent space of a pretrained SD model, semantic grounding enables it to more accurately restore content relevant to downstream segmentation tasks.

\begin{figure}[t]
  \centering
  \includegraphics[width=\columnwidth]{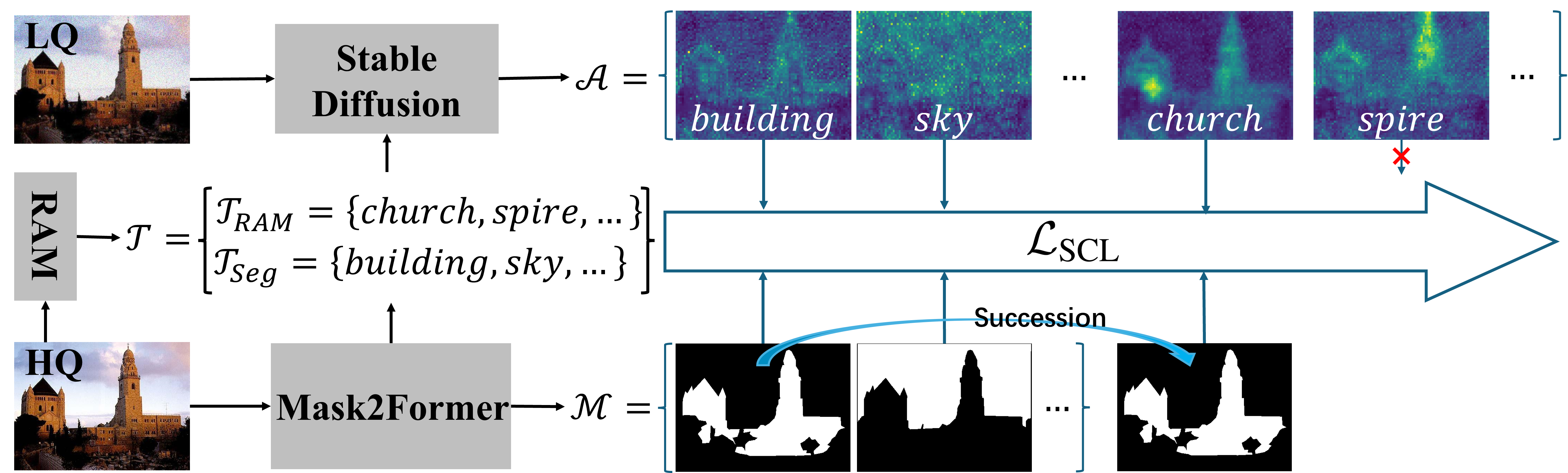}
  \caption{Semantic-Constrained Loss (SCL) computation. Semantically aligned texts inherit corresponding masks (\textit{e.g.}, ``church'' inherits the mask of ``building''), while unmatched terms (\textit{e.g.}, ``spire'') are excluded from SCL.}
  \label{fig:scl}
\end{figure}

Obtaining precise semantic masks is central to SCL. Open-vocabulary models offer broad category coverage but coarse masks, while closed-set models provide finer masks but cover fewer categories. To balance this trade-off, we find that training our model requires only a small amount of high-quality data (see \textbf{Appendix} \ref{apenn2}). We therefore select a subset of semantically rich images based on image quality assessment (IQA) scores and segmentation quality, and combine them with samples from standard segmentation datasets that provide pixel-level annotations. Specifically, we use a pre-trained segmentation model to generate semantic masks \( \mathcal{M} = \{M_1, \dots, M_K\} \) and labels \( \mathcal{T}_{\text{Seg}} = \{T_1, \dots, T_K\} \) from each high-quality image \( \bm{x}_H \). To enrich semantic content, we also use RAM~\cite{zhang2024recognize} to extract additional descriptive labels \( \mathcal{T}_{\text{RAM}} = \{T_1, \dots, T_L\} \) from the same image. The combined label set \( \mathcal{T} = \mathcal{T}_{\text{Seg}} \cup \mathcal{T}_{\text{RAM}} \) is encoded via CLIP’s text encoder to obtain the semantic representation \( c_\mathcal{T} \). Since many tags in \( \mathcal{T}_{\text{RAM}} \) are overlapping or closely related to those in \( \mathcal{T}_{\text{Seg}} \) (\textit{e.g.}, ``people'' vs. ``person''), we manually construct a mapping table (see \textbf{Appendix} \ref{appen3}) to align them, allowing RAM-derived tags to inherit the corresponding masks when applicable (as shown in Figure~\ref{fig:scl}). 

During SCR LoRA training, these text queries \( c_\mathcal{T} \) compute cross-attention with latent visual features from degraded input \( \bm{x}_L \), producing attention maps \( \mathcal{A} = \{A_1, \dots, A_{K+L}\} \), where each \( A_n \) (for \( n = 1, \dots, K+L \)) highlights the spatial region associated with the semantic tag \( T_n \in \mathcal{T} \). These attention maps highlight spatial regions in \( \bm{x}_L \) associated with the semantic concept \( T_n \), guiding the model toward generating visually coherent content. While there is a resolution gap between the typically coarse attention maps and higher-resolution ground-truth masks, we treat these attention maps as structural priors rather than pixel-accurate segmentations: they reliably capture the essential spatial topology of semantic regions across scales. The calculation process of SCL is illustrated in Figure~\ref{fig:scl}. It comprises both region-level and pixel-level components, which can explicitly aligns attention responses with the ground-truth semantic structure, encouraging semantically meaningful focus and enhancing the model’s ability to preserve structural fidelity during restoration. The SCL loss is:
\begin{equation}
\mathcal{L}_{\text{SCL}}\left({A},{M}\right) = \lambda_{\text{region}} \mathcal{L}_{\text{region}} + \lambda_{\text{pixel}} \mathcal{L}_{\text{pixel}} ,
\label{eq:scl}
\end{equation}
\noindent where $\lambda_{\text{region}}$ and $\lambda_{\text{pixel}}$ are two scaling factors. The region-level loss $\mathcal{L}_{\text{region}}$ encourages the attention map \( {A} \in [0,1]^{H \times W} \) to focus its activation within the semantic region ${M} \in \{0,1\}^{H \times W}$, a binary mask where $M_{i,j} = 1$ indicates object presence:
\begin{equation}
    \mathcal{L}_{\text{region}} = 1 - \frac{\sum_{i,j} {A}_{i,j} \cdot {M}_{i,j}}{\sum_{i,j} {A}_{i,j}},
\end{equation}
where $(i,j)$ indexes each spatial location, and ${A}_{i,j} \in [0,1]$ denotes the normalized attention response at that location. This formulation measures the proportion of attention energy that overlaps with the target region, penalizing dispersed or misaligned responses.

To achieve finer semantic consistency, we use the binary cross-entropy loss as the pixel-level loss $\mathcal{L}_{\text{pixel}}$:
\begin{equation}
\begin{split}
\mathcal{L}_{\text{pixel}} = -\frac{1}{HW} \sum_{i=1}^{H} \sum_{j=1}^{W} \Bigl[ & {M}_{i,j} \log({A}_{i,j}) \\
& + (1 - {M}_{i,j}) \log(1 - {A}_{i,j}) \Bigr].
\end{split}
\end{equation}
It is important to note that the lengths of \( \mathcal{M} \) and \( \mathcal{A} \) are not equal. When computing \( \mathcal{L}_{\text{SCL}} \), each pair of \( M \in \mathcal{M} \) and \( A \in \mathcal{A} \) is matched based on the associated tag \( T \). Attention maps \( A \) without a corresponding semantic mask \( M \) are excluded from the loss computation. In summary, SCL is implemented as a multi-resolution-layer constraint that aligns coarse attention responses from multiple transformer layers with the corresponding semantic masks; this provides explicit structural regularization during training and helps to suppress spurious activations across layers.

The total objective for the SCR model is the direct sum of the restoration loss and the semantic-constrained loss, as the specific weighting is handled internally within the SCL components ($\lambda_{\text{region}}$ and $\lambda_{\text{pixel}}$): 
\begin{equation}
    \mathcal{L}_{\text{total}} = \mathcal{L}_{\text{res}} + \mathcal{L}_{\text{SCL}}.
\end{equation}

\subsection{Restoration Adaptation Segmentation Model}
High-quality visual representations are crucial for achieving accurate semantic segmentation, especially when the input images suffer from severe degradations. To this end, we first conduct a restoration-focused pretraining stage, where the model learns to reconstruct clean images from their degraded counterparts. This restoration stage significantly enhances the perceptual quality and structural consistency of the intermediate features, which are essential for downstream dense prediction tasks such as segmentation.

To effectively transfer restoration knowledge to the segmentation task, we propose a Restoration Adaptation Segmentation (RAS) model based on the LoRA framework. Specifically, the SCR LoRA parameters \(\Delta \theta_{scr}\), trained during the restoration stage to handle various degradation types, are integrated into the backbone network as its initialization. Consequently, the latent diffusion network is initialized with \( \epsilon_{\theta_{scr}} \), replacing the original \( \epsilon_\phi \). On top of this restored backbone, we introduce an Seg LoRA branch dedicated to segmentation, denoted as \(\Delta \theta_{seg}\), which is fine-tuned specifically for the downstream segmentation task, as illustrated in Figure~\ref{fig3}. The complete parameter set of RASS is therefore defined as \(\theta_{rass} = \{\phi, \Delta \theta_{scr}, \Delta \theta_{seg}\}\). This modular design enables the seamless transfer of degradation-invariant features learned during restoration to the segmentation stage, thereby enhancing the model's capacity to extract semantically meaningful representations from LQ inputs.

Unlike restoration models, segmentation models do not require the \( \epsilon_\phi \) to predict noise \( \hat{\epsilon} \), nor do they rely on the \( D_\phi \) to reconstruct the HQ image from the refined latent representation. Instead, existing segmentation frameworks typically depend on multi-scale feature representations to achieve accurate pixel-wise classification. To this end, during the Seg LoRA training phase, we replace the \( D_\phi \) with a trainable segmentation head, denoted as \( S_\phi \), which takes as input the hierarchical features \( \mathcal{F}_\epsilon = \{ f^{(i)}_\epsilon \}_{i=1}^{N_\epsilon} \) extracted from the latent diffusion network and predicts the corresponding segmentation mask. In addition, we incorporate features \( \mathcal{F}_E = \{ f^{(i)}_E \}_{i=1}^{N_E} \) extracted from \( E_\phi \) into \( S_\phi \). This is because the latent space is inherently compressed and may lose important fine-grained details, while the encoder features can help retain spatial precision and structural information crucial for accurate segmentation. To supervise this process, we adopt a compound segmentation loss $\mathcal{L}_{\text{seg}}$ composed of classification and mask-matching objectives: 
\begin{equation}
    \mathcal{L}_{\text{seg}}\left(\hat{\bm{y}}, \bm{y} \right) = \lambda_{\text{CE}} \mathcal{L}_{\text{CE}} + \lambda_{\text{Dice}} \mathcal{L}_{\text{Dice}} + \lambda_\text{{cls}} \mathcal{L}_{\text{CE}}^{\text{cls}},
\end{equation}

\noindent where \( \hat{\bm{y}} = S_{\phi}(\mathcal{F}_\epsilon , \mathcal{F}_E) \) denotes the predicted segmentation mask, \(\bm{y}\) is the ground-truth segmentation mask, $\mathcal{L}_{\text{CE}}$ is the binary cross-entropy loss applied to predicted masks, and $\mathcal{L}_{\text{Dice}}$ is the soft Dice loss~\cite{milletari2016v} promoting spatial overlap with ground-truth masks. The term $\mathcal{L}_{\text{CE}}^{\text{cls}}$ denotes classification loss used for category prediction. Following settings in \cite{cheng2022masked}, we set $\lambda_{\text{CE}} = \lambda_{\text{Dice}} = 5.0$ and $\lambda_{\text{cls}} = 2.0$. For unmatched predictions, a reduced weight $\lambda_{\text{cls}} = 0.1$ is applied to mitigate overfitting on background categories.

\section{Experiments}

\subsection{Experimental Settings}
\label{sec:exp_set}
\noindent
\textbf{Training and Testing Datasets.} For RASS, we use ADE20K \cite{zhou2017ade20k} (150 categories) for training and evaluation, applying the Real-ESRGAN \cite{wang2021real} degradation pipeline to simulate LQ inputs. We also construct a real-world LQ dataset, namely RealLQ, containing 100 degraded images from RealSR \cite{cai2019toward}, DrealSR \cite{wei2020component}, and SR-Raw \cite{zhang2019zoom}, covering various scenes with 82 annotated categories. This dataset is augmented with synthetic LQ data for comprehensive evaluation, and detailed settings are provided in the \textbf{Appendix} \ref{apenn1}. To improve the effectiveness of SCR in RASS, we train on 469 ADE20K images together with 475 curated LSDIR \cite{li2023lsdir} images, using the same degradation pipeline to synthesize LR-HR pairs. Evaluation is conducted on 3,000 synthetic LQ images from DIV2K-Val \cite{agustsson2017ntire} and real LQ-HQ pairs from DrealSR and RealSR.

\begin{figure*}[t]
	\centering
	\includegraphics[width=\textwidth]{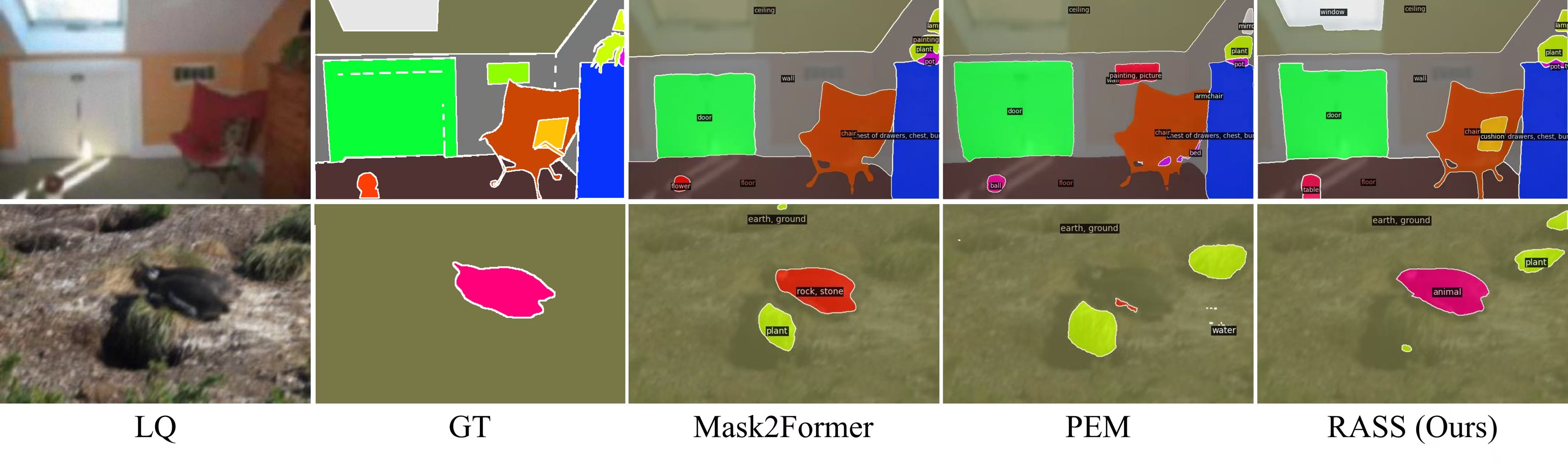}
    \vspace{-5mm}
	\caption{Comparison of segmentation results of different methods, the above samples are from the simulated degradation ADE20K (first row) and RealLQ (second row) dataset. The comparison models are fine-tuned with LQ images.}
	\label{fig:seg_example}
    % \vspace{-0.3cm}
\end{figure*}

\noindent
\textbf{Implementation Details.} We train the SCR model following the settings of OSEDiff \cite{wu2024one}, using 8 NVIDIA H20 GPUs for about 1 day with a batch size of 16. The AdamW optimizer \cite{loshchilov2017adamw} is adopted with a learning rate of 5e-5. The model for obtaining semantic masks is Mask2Former \cite{cheng2022masked}. The RAS model uses Mask2Former \cite{cheng2022masked} as the segmentation head and follows its settings for 160,000 iterations on the same hardware setup, using AdamW with an initial learning rate of 1e-4 and weight decay of 5e-2. 
During inference, $RASS_{base}$ replaces Mask2Former to extract category labels \( \mathcal{T}_{\text{Seg}} \). Note that $RASS_{base}$ serves as the baseline version of our RASS, utilizing OSEDiff\textsuperscript{+} (with a frozen VAE encoder) as the restoration adaptation module. Additionally, following the protocol of OSEDiff~\cite{wu2024one}, we use DAPE \cite{wu2024seesr} instead of RAM to generate \( \mathcal{T}_{\text{RAM}} \).

\noindent
\textbf{Evaluation Metrics.} We use mean Intersection over Union (mIoU), a standard metric that measures the overlap between predicted and ground-truth segmentation masks, to evaluate segmentation performance. Frames-per-second (FPS) is measured on an NVIDIA H20 GPU, with a batch size of 1 by taking the average runtime on the entire validation set, including post-processing time. For image quality, we adopt both full-reference and no-reference metrics. PSNR and SSIM \cite{ssim} measure pixel-level fidelity, while DISTS \cite{dists} assesses perceptual quality. NIQE \cite{niqe}, MANIQA \cite{maniqa}, MUSIQ \cite{musiq}, and CLIPIQA \cite{maniqa} are used as no-reference metrics to evaluate overall image quality.

\noindent
\textbf{Compared Methods.} 
For the segmentation task, due to the limited availability of public implementations for LQ image segmentation models and their significantly different performance compared to standard-quality models, we compare with several SOTA segmentation models on standard-quality images, including SegFormer-B5~\cite{xie2021segformer}, Mask2Former-L~\cite{cheng2022masked}, FeedFormer-B2~\cite{shim2023feedformer}, SeaFormer-L~\cite{wan2023seaformer}, and PEM-L\textsuperscript{+}~\cite{cavagnero2024pem} (enhanced version using swin-L as backbone). All models are fine-tuned on the same training data using their official code and checkpoints. For the generative restoration task, we evaluate SCR against SOTA Real-IR methods, including StableSR~\cite{wang2023exploiting}, SeeSR~\cite{wu2024seesr}, DiffBIR~\cite{lin2023diffbir}, OSEDiff~\cite{wu2024one}, and PiSA-SR~\cite{sun2025pixel}, all of which are based on diffusion models. Publicly released codes and models are used for evaluation.

\subsection{Comparison with State-of-the-Arts}
\label{sec:exp:comp}
% \vspace{-5pt}
% \noindent
\textbf{Quantitative Comparisons.}
\label{sec:exp:qc}
To enable a fair comparison between RASS and SOTA segmentation models on LQ images, we adopt three evaluation protocols. First, in Direct Testing (DT), models trained on SQ data are directly evaluated on LQ images. Second, in Restoration-to-Segmentation (R2S), LQ images are first restored using SCR (chosen for its strong semantic preservation) and then fed into the same models as used in DT. Third, in Fine-Tuning (FT), models originally trained on SQ data are further fine-tuned using LQ data before being tested on LQ images. Since RASS is trained on LQ data, its DT results are not applicable and thus omitted (denoted by `-' in Table~\ref{tab:lq}). However, since it generalizes well to HQ inputs, its R2S results are reported. In FT, although RASS is not pre-trained on SQ, all models are ultimately trained on LQ data, ensuring a fair comparison.

\begin{table*}[htbp]
\caption{Semantic segmentation (mIoU↑) on degraded ADE20K-val and RealLQ. Symbol `\textsuperscript{+}' denotes an enhanced version, and `-' denotes not applicable. Best results are highlighted in \textbf{bold}.}
\label{tab:lq}
\centering
\small
\renewcommand{\arraystretch}{1.1}
\resizebox{0.9\textwidth}{!}{
\begin{tabular}{l|ccc|ccc|ccc}
\toprule
\multirow{2}{*}[-0.6ex]{Method} & \multicolumn{3}{c|}{ADE20K} & \multicolumn{3}{c|}{RealLQ} & \multirow{2}{*}[-0.6ex]{FPS} & \multirow{2}{*}[-0.6ex]{Params} & \multirow{2}{*}[-0.6ex]{\makecell{Params \\ Trainable}}
 \\
\cmidrule(lr){2-4} \cmidrule(l){5-7}
 & DT & R2S & FT & DT & R2S & FT & & & \\
\midrule
SegFormer~\cite{xie2021segformer} & 26.37 & 44.34 & 34.57 & 22.54 & 25.31 & 22.63 & - & 84.7M & 84.7M \\
Mask2Former~\cite{cheng2022masked} & 37.15 & 45.64 & 44.19 & 33.15 & 34.83 &  34.41 &  5.4 & 215.5M & 215.5M \\
FeedFormer~\cite{shim2023feedformer} & 22.62 & 41.39 & 23.98 & 20.87 & 21.19 & 19.19 & \textbf{50.4} & 29.1M & 27.9M \\
SeaFormer~\cite{wan2023seaformer} & 16.39 & 38.09 & 23.24 & 15.52 & 17.86 &  14.97 & 31.1 & \textbf{14M} & \textbf{14M} \\
PEM \textsuperscript{+}~\cite{cavagnero2024pem} & 33.55 & 45.54 & 41.52 & 33.52 & 31.80 & 34.38 & 9.7 & 207.1M & 207.1M \\
% \addlinespace[0.5ex]\hdashline
RASS (Ours) & - & \textbf{46.81} & \textbf{47.42} & - & \textbf{37.84} &  \textbf{39.80}  &  3.4 & 1788.7M &  27.9M\\
\bottomrule
\end{tabular}
}
\end{table*}

Table~\ref{tab:lq} summarizes the comparison results, RealLQ shares the same testing resolution and FPS as ADE20K. SegFormer was not benchmarked for FPS in this specific environment due to version limitations and is marked as ``-''. All baseline models suffer from significant performance drops under the DT setting, reflecting limited robustness to LQ inputs. In contrast, our RASS achieves the best performance (47.42 mIoU), outperforming the strongest baseline Mask2Former (44.19 mIoU with FT) by 3.23 mIoU. While RASS (1.7B params) operates at 3.4 FPS—slower than Mask2Former (5.4 FPS)—this computational cost is justified by the necessity of capacity: efficiency-focused models like SeaFormer (14M params) fail to handle degradations (only 14.97 mIoU). In addition, although RASS leverages a much larger backbone, our proposed Restoration Adaptation Strategy enables training only 27.9M parameters, which is significantly fewer than Mask2Former’s 215.5M trainable parameters. Importantly, at inference RASS performs only one-step denoising in latent space: the input is encoded by the VAE encoder and denoised by the U-Net, and the resulting cleaned latent features are fed directly to the segmentation head. This avoids running the computationally expensive VAE decoder and reduces inference cost compared to full image reconstruction. Under the R2S protocol, most baselines benefit from enhanced inputs, but RASS remains the top performer (46.81 mIoU). Although baselines improve moderately with fine-tuning, they still lag behind RASS. On the RealLQ dataset, FT does not consistently outperform DT; lightweight models like FeedFormer and SeaFormer even degrade after fine-tuning, possibly due to the domain gap between synthetic degradations in ADE20K and real-world LQ characteristics. Restoration generally improves performance under R2S, except for PEM\textsuperscript{+}, which experiences a 1.82 mIoU drop—indicating that restoration may sometimes be detrimental due to residual quality gaps. This highlights the advantage of RASS, which remains effective without relying on the quality of preprocessing. Although Figure \ref{fig:seg_example} shows that RASS performs well, some pixels are still missegmented—for example, the target in the lower-left corner of the first row. This may result from excessive compression of small targets in the latent space, making them harder to recognize.

\begin{table*}[htbp]
\caption{Quantitative comparison with Diffusion-based methods on DrealSR, RealSR and DIV2K datasets. The suffix -N indicates the number of inference steps (\textit{e.g.}, -200 for 200 steps, -1 for single-step), which affects the inference time and efficiency. Best results are highlighted in \textbf{bold}.}

\label{tab:diff}
\centering
\small
\renewcommand{\arraystretch}{1.05}
\resizebox{\textwidth}{!}{
\begin{tabular}{l|l|cccccccc}
\toprule
Dataset & Method & PSNR↑ & SSIM↑ & DISTS↓ & NIQE↓ & MUSIQ↑ & MANIQA↑ & CLIPIQA↑ \\
\midrule
\multirow{6}{*}[0pt]{\centering DrealSR}
& StableSR-200~\cite{wang2023exploiting} & 28.03 & 0.7536 & 0.2269 & 6.5239 & 58.51 & 0.5593 & 0.6356 \\
& DiffBIR-50~\cite{lin2023diffbir} & 26.71 & 0.6571 & 0.2748 & 6.3124 & 61.07 & 0.5930 & 0.6395 \\
& SeeSR-50~\cite{wu2024seesr} & 28.14 & 0.7712 & 0.2299 & 6.4523 & 64.76 & 0.6005 & 0.6897 \\
& OSEDiff-1~\cite{wu2024one} & 27.92 & \textbf{0.7835} & \textbf{0.2165} & 6.4902 & 64.65 & 0.5899 & 0.6963 \\
& PiSA-SR-1~\cite{sun2025pixel} & \textbf{28.31} & 0.7804 & 0.2169 & \textbf{6.1837} & 66.10 & 0.6146 & 0.6968 \\
& SCR-1 (Ours) & 28.16 & 0.7743 & 0.2302 & 6.4616 & \textbf{67.16} & \textbf{0.6394} & \textbf{0.7056} \\
\midrule
\multirow{6}{*}[0pt]{\centering RealSR}
& StableSR-200~\cite{wang2023exploiting} & 24.65 & 0.7080 & 0.2140 & 5.8809 & 65.88 & 0.6227 & 0.6233 \\
& DiffBIR-50~\cite{lin2023diffbir} & 24.75 & 0.6567 & 0.2312 & 5.5346 & 64.98 & 0.6243 & 0.6463 \\
& SeeSR-50~\cite{wu2024seesr} & 25.21 & 0.7216 & 0.2218 & 5.3970 & 69.70 & 0.6428 & 0.6673 \\
%& ResShift-S15 & 25.6647 & 0.7360 & 0.3279 & 0.2475 & 8.0722 & 56.8920 & 0.5089 & 0.5361 \\
%& SinSR-S1 & \textbf{26.2606} & 0.7345 & 0.3218 & 0.2357 & 6.3445 & 60.3050 & 0.5379 & 0.6149 \\
& OSEDiff-1~\cite{wu2024one} & 25.15 & 0.7341 & 0.2128 & 5.6521 & 69.08 & 0.6326 & 0.6698 \\
& PiSA-SR-1~\cite{sun2025pixel} & \textbf{25.50} & \textbf{0.7418} & \textbf{0.2044} & 5.5054 & 70.15 & 0.6552 & 0.6698 \\
& SCR-1 (Ours) & 25.34 & 0.7219 & 0.2149 & \textbf{5.2387} & \textbf{70.27} & \textbf{0.6689} & \textbf{0.6718} \\
\midrule
\multirow{6}{*}[0pt]{\centering DIV2K}
& StableSR-200~\cite{wang2023exploiting} & 23.26 & 0.5726 & 0.2048 & 4.7581 & 65.92 & 0.6188 & 0.6771 \\
& DiffBIR-50~\cite{lin2023diffbir} & 23.64 & 0.5647 & 0.2128 & 4.7042 & 65.81 & 0.6210 & 0.6704 \\
& SeeSR-50~\cite{wu2024seesr} & 23.73 & 0.6056 & 0.1966 & 4.7903 & 68.42 & 0.6220 & 0.6867 \\
%& ResShift-S15 & \textbf{24.5931} & \textbf{0.6232} & 0.3077 & 0.2136 & 6.9156 & 58.8998 & 0.5279 & 0.5715 \\
%& SinSR-S1 & 24.4115 & 0.6018 & 0.3239 & 0.2064 & 5.9973 & 62.8036 & 0.5393 & 0.6494 \\
& OSEDiff-1~\cite{wu2024one} & 23.72 & \textbf{0.6109} & 0.1975 & 4.7108 & 67.96 & 0.6147 & 0.6680 \\
& PiSA-SR-1~\cite{sun2025pixel} & \textbf{23.86} & 0.6058 & \textbf{0.1934} & \textbf{4.5555} & 69.67 & 0.6401 & 0.6928 \\
& SCR-1 (Ours) & 23.75 & 0.6029 & 0.2231 & 4.5793 & \textbf{70.02} & \textbf{0.6561} & \textbf{0.6971} \\
\bottomrule
\end{tabular}
}
% \vspace{-0.2cm}
\end{table*}

\begin{figure}[t]
	\centering
	\includegraphics[width=\textwidth]{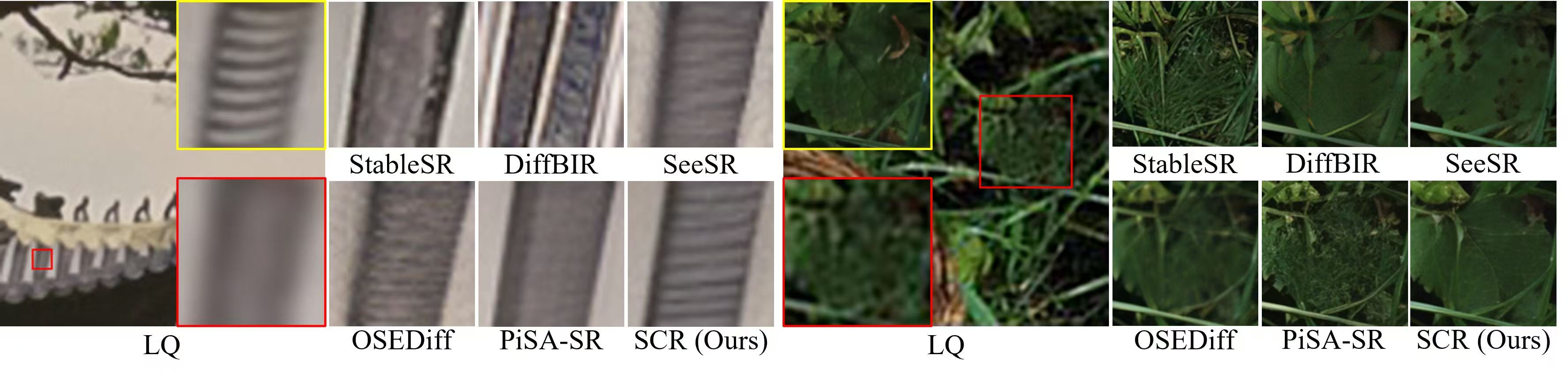}
	\caption{Comparison of restoration results of different Real-IR methods, the above samples are from RealSR (left) and DIV2K (right) datasets. Red boxes highlight zoom-in areas; yellow boxes indicate ground truth. Please zoom in for clarity.}
	\label{fig:res_example}
    \vspace{-0.3cm}
\end{figure}

To demonstrate the effectiveness of our SCR, we conduct a comprehensive performance comparison with other SOTA models on the common RealIR test benchmark. It is worth noting that since our SCR adopts the OSEDiff framework and is trained on the selected dataset with semantic constraints, it inherently supports one-step denoising. As illustrated in Table~\ref{tab:diff}, SCR demonstrates superior no-reference perceptual quality, achieving the highest MUSIQ, MANIQA, and CLIPIQA scores on both DrealSR, RealSR and DIV2K, and the best NIQE performance on RealSR while remaining competitive on DrealSR and DIV2K. Furthermore, despite using only a single sampling step and being trained on limited data, SCR attains competitive full-reference performance, ranking among the top in PSNR and SSIM. It outperforms most single-step baselines and approaches multi-step counterparts such as StableSR and SeeSR, effectively balancing perceptual quality and fidelity. The visualization results in Figure \ref{fig:res_example} show that SCR not only has strong semantic understanding capabilities (recovering the degraded leaf), providing a foundation for RASS, but also surpasses other models in restoring fine texture details (recovering the tiles on the roof).

Additional visual analysis, quantitative results comparing RASS with refinement-based methods, and quantitative results comparing SCR with non-diffusion baselines (\textit{e.g.}, GAN-based methods) are provided in the \textbf{Appendix} \ref{appen4} and \ref{appen5}.
% \begin{figure}[t]
%   \centering
%   \includegraphics[width=\columnwidth]{image/res_out_radar.png}
%   \caption{Performance comparison on the RealSR (left) and DIV2K (right) datasets. The suffix `-N' denotes the number of inference steps (\textit{e.g.}, `-200' indicates 200 steps; `-1' denotes a single-step inference). Metrics marked with `$\uparrow$' indicate that higher values are better, while those marked with `$\downarrow$' indicate that lower values are preferable.}
%   \label{fig:res}
%   %\vspace{-0.3cm}
% \end{figure}

\subsection{Ablation Studies}
\label{sec:exp:ab}
We conducted ablation studies to validate the contribution of each component in RASS. Additional results, including hyperparameter analysis of SCL, are provided in the \textbf{Appendix} \ref{apenn2}.

\begin{table}[htbp]
\centering
\caption{Segmentation and restoration results on ADE20K-val under simulated degradation. Symbol `\textsuperscript{+}' denotes an enhanced version. Best results are highlighted in \textbf{bold}.}
\label{tab:scr_ab}
\renewcommand{\arraystretch}{1.1}
\begin{tabular}{lccccc}
\toprule
Method & mIoU↑ & LPIPS↓ & DISTS↓ & MUSIQ↑ & MANIQA↑ \\
\midrule
OSEDiff & 46.05 & 0.2731 & 0.1625 & 62.95 & 0.6506 \\
OSEDiff \textsuperscript{+} & 46.24 & 0.2653 & 0.1613 & 64.93 & 0.6702 \\
SCR (Ours) & \textbf{46.81} & \textbf{0.2628} & \textbf{0.1595} & \textbf{65.46} & \textbf{0.6746} \\
\bottomrule
\end{tabular}
\end{table}

\textbf{Effectiveness of Semantic Restoration} in Table~\ref{tab:scr_ab}. To assess semantic restoration, we compare the restoration results of OSEDiff, OSEDiff\textsuperscript{+} (inspired by PiSA-SR~\cite{sun2025pixel} with a frozen VAE encoder), and our SCR model on the degraded ADE20K dataset, and their performance on downstream segmentation tasks (R2S setting), with RASS as the segmentation model, as it achieves the best performance in R2S in Table~\ref{tab:lq}. SCR introduces a semantically constrained loss aligned with segmentation priors and achieves the best performance, indicating improved visual and semantic consistency. 

\begin{table}[htbp]
\centering
\small
\caption{Ablation results of RAS on the restoration adaptations. All experiments are tested on degraded ADE20K dataset. Symbol \textsuperscript{+} denotes an enhanced version. Best results are highlighted in \textbf{bold}.}
\label{tab:seg_ab}
\begin{tabular}{lc}
\toprule
Restoration Adaptation & mIoU $\uparrow$ \\
\midrule
None & 44.56 \\
OSEDiff\textsuperscript{+} & 47.05 \\
SCR (Ours) & \textbf{47.42} \\
\bottomrule
\end{tabular}
\end{table}

\textbf{Effectiveness of Restoration Adaptation} in Table~\ref{tab:seg_ab}. To evaluate restoration adaptation, we test RASS with three variants: no restoration prior (None), OSEDiff\textsuperscript{+}, and SCR, all using LQ images as input. It's worth noting that in the ``None'' setting, we omit the first stage SCR model (restoration adaptation) and directly train the RAS model on the pre-trained T2I backbone. RASS consistently outperforms the others, confirming the benefit of integrating restoration into segmentation. 

\begin{table}[htbp]
\caption{Ablation results of SCR model on the additional description (AD). All experiments are tested on the DrealSR dataset.}
\label{tab:wo_ab}
\centering
\small
\renewcommand{\arraystretch}{1.05}
\begin{tabular}{lcccccc}
\toprule
Train & Testing  & LPIPS↓ & DISTS↓ & MUSIQ↑ & MANIQA↑ \\
\midrule
w/o AD & w/o AD & 0.3143 & 0.2365 & 66.82 & 0.6338 \\
w/ AD & w/o AD & 0.3094 & 0.2302 & 67.15 & 0.6393 \\
w/ AD & w/ AD & 0.3094 & 0.2302 & 67.16 & 0.6394 \\
\bottomrule
\end{tabular}
\end{table}

\textbf{Effectiveness of Additional Descriptive} in Table~\ref{tab:wo_ab}. As stated in Section~3.2, we use RAM as an optional module to extract additional descriptive labels during training, while DAPE is employed at the inference stage. To evaluate the impact of descriptive inputs, we compare models trained using only category names with those trained using RAM-generated descriptions. While incorporating additional descriptions (w/ AD) leads to consistent performance improvements, the model remains stable even without such descriptions (w/o AD) during training. Moreover, enabling or disabling AD at inference results in only marginal performance variations. Based on these observations, we adopt additional descriptions by default.

\begin{table}[htbp]
\caption{Ablation results of RAS model on the VAE features. All experiments are tested on the degraded ADE20K dataset.}
\label{tab:fe}
\centering
\small
\renewcommand{\arraystretch}{1.05}
\resizebox{0.9\textwidth}{!}{
\begin{tabular}{cccccc}
\toprule
UNet Features & VAE Encoder Features & VAE Decoder Features & mIoU & FPS & Params \\
\midrule
$\surd$ & $\times$ & $\times$ & 46.85 & 3.8 & 1788.1M \\
$\surd$ & $\surd$ & $\times$ & 47.42 & 3.4 & 1788.7M \\
$\surd$ & $\times$ & $\surd$ & 47.25 & 2.5 & 1788.6M \\
$\surd$ & $\surd$ & $\surd$ & 47.60 & 2.4 & 1789.3M \\
\bottomrule
\end{tabular}
}
\end{table}

\textbf{Effectiveness of VAE Features}. Table \ref{tab:fe} demonstrates the impact of VAE features on segmentation.  While decoder features are beneficial for segmentation, the decoder-side option introduces a significant runtime overhead: the inference speed drops from 3.4 FPS to 2.5 FPS (approximately a 30\% reduction), which is mainly due to the computationally intensive upsampling operations involved in the VAE decoder. Considering the trade-off between performance and efficiency, we do not use decoder features by default.

\section{Discussion}
Despite the promising results, the large parameter size of our RASS model (due to its use of the pre-trained SD model) poses challenges for efficient inference. In addition, small object segmentation remains challenging, as the SD training strategy compresses features into the latent space via VAE, often discarding critical details essential for identifying small targets. Future work will focus on enhancing efficiency and improving small object detection capabilities. We plan to explore lightweight unified architectures to reduce inference costs. Additionally, we will investigate multi-scale feature reconstruction and high-resolution refinement strategies to better preserve fine details and improve segmentation performance for small objects.

\section{Conclusion}\label{sec13}
We proposed RASS, a unified framework designed to improve the semantic segmentation performance of LQ images, which bridged image restoration and segmentation through LoRA-based module merging and task-specific fine-tuning. By incorporating semantic priors into the image restoration process, our unified framework achieved semantically meaningful and visually consistent reconstructions that benefit downstream segmentation. To support real-world applicability, we constructed a high-quality annotated dataset of real LQ images. Extensive experiments on both synthetic and real LQ benchmarks showed that our method achieved SOTA performance in both restoration and segmentation, validating the effectiveness of RASS.

\section{Data Availability}
The data that support the findings of this study are openly available at the following links corresponding to each dataset: \url{https://ade20k.csail.mit.edu/} (ADE20k), \url{https://ofsoundof.github.io/lsdir-data/} (LSDIR), \url{https://github.com/csjcai/RealSR} (RealSR), \url{https://github.com/xiezw5/Component-Divide-and-Conquer-for-Real-World-Image-Super-Resolution} (DrealSR), and \url{https://data.vision.ee.ethz.ch/cvl/DIV2K/} (DIV2K). All datasets are accessed in accordance with their respective official usage licenses, and no restricted access or additional permissions are required for academic research purposes as specified by each dataset's owner. The RealLQ dataset can be obtained through \url{https://github.com/Ka1Guan/RASS.git}.

\bibliographystyle{unsrtnat}
\bibliography{rass}

\section{Appendices} \label{appen}
In the file, we provide the following materials:

\begin{itemize}
    \item \textit{RealLQ Dataset Setup.} More details of the collection of RealLQ.
    
    \item \textit{Mapping Table Setup.} More details of the mapping table.
    
    \item \textit{Ablation Studies.} Ablation experiments of SCR training and hyperparameter selection on SCL.

    \item \textit{Visualization Results.} Cross-attention map visualization analysis and visual comparisons.
    
\end{itemize}

% \subsection{Real-world Low-quality Image Dataset for Segmentation}

\begin{figure*}[htbp]
	\centering
	\includegraphics[width=\textwidth]{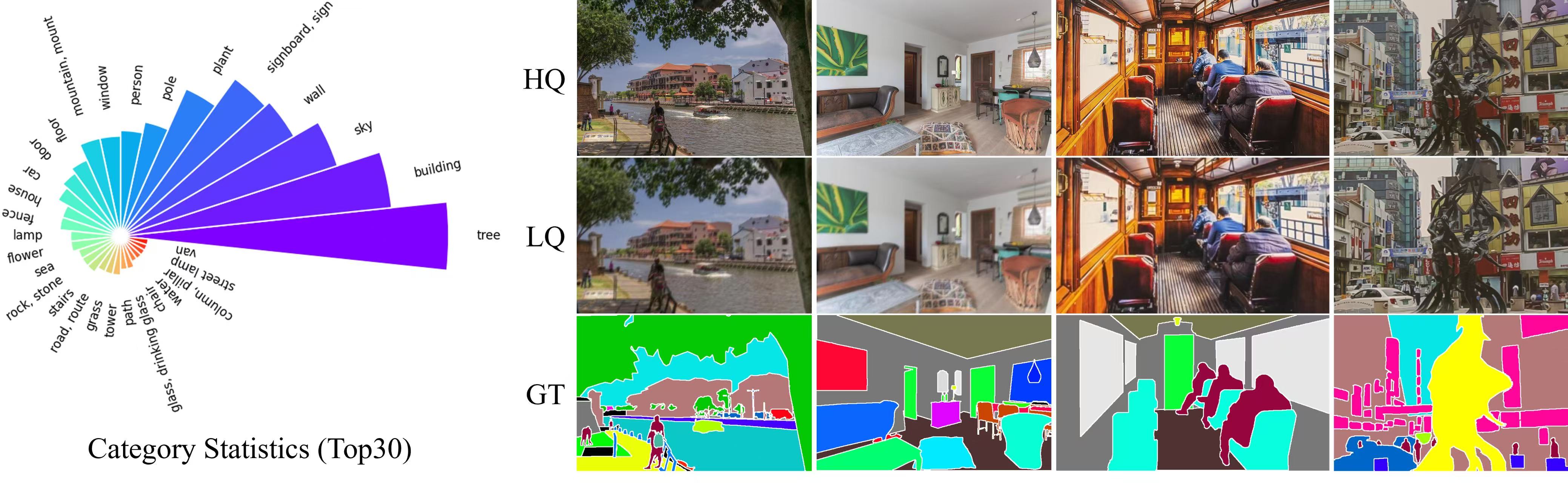}
	\caption{RealLQ category statistics and some sample displays.}
	\label{fig:reallq}
    % \vspace{-0.3cm}
\end{figure*}

\subsection{RealLQ Dataset Setup} \label{apenn1}
% \textbf{RealLQ Dataset.}
To evaluate the generalization capability of the proposed RASS on real-world low-quality images, we introduce a new dataset named RealLQ. Benefiting from the outstanding work of RealSR \cite{cai2019toward}, DRealSR \cite{wei2020component}, and SR-Raw \cite{zhang2019zoom}, we collect 100 LQ-HQ image pairs from these datasets. Pixel-level semantic annotations are conducted on the HQ images using an annotation assistance tool \textit{`ISAT'} \cite{ISAT_with_segment_anything} and manually fine-tuned, ensuring that the semantic labels are well aligned with the corresponding objects in the LQ images. The annotation categories follow the 150-class ADE20K taxonomy (100 ``things'' and 50 ``stuff''), and across the 100 images. A total of 82 distinct categories are annotated, covering both indoor and outdoor scenes. All LQ images are resized to a short side of 512 pixels, with the segmentation maps correspondingly aligned. As a dedicated benchmark for real-world low-quality image segmentation, RealLQ helps bridge the gap in this underexplored area. Dataset samples and statistics are shown in Figure~\ref{fig:reallq}.

\subsection{Ablation Study} \label{apenn2}
\textbf{SCR Training Data.} Through our experiments, we observe that only a small number of samples are required to effectively train the restoration model. As shown in Table~\ref{tab:ss}, we evaluate the baseline restoration model (an improved version of OSEDiff with the VAE encoder frozen) using different amounts of training data. ``All'' denotes the full LSDIR training set containing 84,991 images. We then randomly sample three subsets of 10,000, 2,000, and 1,000 images. The results show that, given the same training time, model performance remains relatively stable even with significantly reduced training data.

\begin{table}[htbp]
\caption{Ablation results of SCR model on training data. All experiments are tested on the DrealSR dataset. `Size' indicates the amount of training data from LSDIR (\textit{e.g.}, `all' for all 84991 images, `475' for 475 images), except for 518, which are from ADE20K.}

\label{tab:ss}
\centering
\small
\renewcommand{\arraystretch}{1.05}
\begin{tabular}{l|cccccccc}
\toprule
Size & PSNR↑ & SSIM↑ & DISTS↓ & NIQE↓ & MUSIQ↑ & MANIQA↑ & CLIPIQA↑ \\
\midrule

All & 28.54 & 0.7932 & 0.2113 & 6.5518 & 64.51 & 0.5987 & 0.6730 \\
10000 & 28.47 & 0.7868 & 0.2157 & 6.6124 & 64.95 & 0.6032 & 0.6761 \\
2000 & 28.52 & 0.7874 & 0.2112 & 6.2926 & 64.86 & 0.6013 & 0.6683 \\
1000 & 28.45 & 0.7852 & 0.2185 & 6.3278 & 64.17 & 0.5975 & 0.6700 \\
475 & 27.43 & 0.7303 & 0.2457 & 5.9606 & 69.12 & 0.6569 & 0.7018 \\
475+518 & 28.14 & 0.7762 & 0.2259 & 6.4879 & 66.73 & 0.6274 & 0.6883 \\
\bottomrule
\end{tabular}
\end{table}

Since high-quality images are more beneficial for training restoration models, and our SCR training data needs to provide sufficiently good closed-set segmentation masks (to facilitate downstream semantic segmentation tasks), we select 475 high-quality samples based on image quality assessment (IQA) scores and segmentation quality. Specifically, we use the \textit{maniqa-pipal} and \textit{topiq\_nr-flive} functions in \textit{`PYIQA Toolbox'} \cite{pyiqa} to identify the 1,000 images with the highest MANIQA \cite{maniqa} and TOPIQA \cite{chen2024topiq} scores from the entire LSDIR dataset. We then segment each image using Mask2Former to obtain a semantic segmentation map. Based on the completeness and richness of the segmentation results, we manually select 475 training images (with an average score of 0.7697 for MANIQA and 0.7632 for TOPIQA). Training with this subset yields strong results in no-reference metrics (\textit{e.g.}, NIQE, MUSIQ, MANIQA, CLIPIQA), but shows a notable drop in full-reference metrics (\textit{e.g.}, PSNR, SSIM, DISTS). This decline likely stems from limited generalization due to the reduced data size and the fact that full-reference metrics rely more heavily on pixel-level similarity to the ground truth. In SD-based generative restoration, however, the ground-truth image is not always the clearest or most detailed, yet restored outputs are still expected to align stylistically with it.

To balance no-reference and full-reference performance and benefit downstream segmentation tasks, we augment the training set by introducing 518 images from the ADE20K dataset (also filtered based on MANIQA and TOPIQA), enhanced using a restoration model (OSEDiff) to improve their IQA scores (averaging 0.7147 MANIQA and 0.7536 TOPIQA). These images include precise human-annotated segmentation masks, which benefit SCR training. Ultimately, our training set consists of 475 LSDIR images and 518 enhanced ADE20K images. As shown in the results for the sample size of 475+518 in Table~\ref{tab:ss}, this setting preserves strong no-reference metric performance while improving full-reference scores.

\begin{figure}[htbp]
	\centering
	\includegraphics[width=0.8\linewidth]{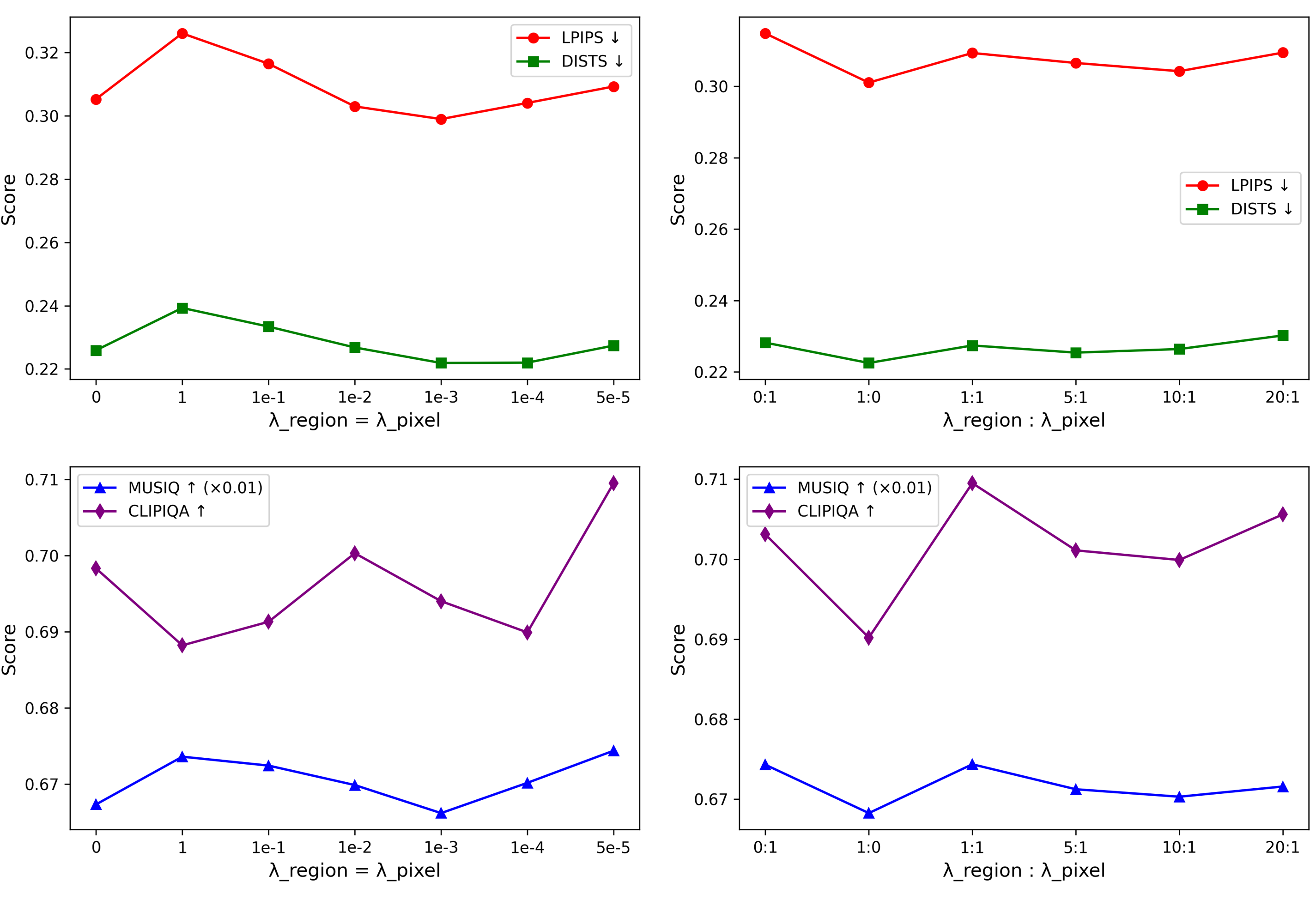}
	\caption{Comparison of different $\lambda_{\text{region}}$ and $\lambda_{\text{pixel}}$ settings. The quantization results from the DrealSR dataset.}
	\label{fig:lamda_ab}
    % \vspace{-0.3cm}
\end{figure}

\textbf{Hyperparameters in Semantic-Constrained Loss.} 
We perform ablation experiments on the two scaling factors, $\lambda_{\text{region}}$ and $\lambda_{\text{pixel}}$, in Equation (2). We begin by setting both factors to the same value to assess the overall impact of scaling on model performance. As shown in the two line graphs on the left of Figure~\ref{fig:lamda_ab} (where 0 denotes the baseline model without the additional loss), the results suggest that smaller scaling factors are more favorable for training the restoration model. Building on this observation, we further investigate the balance between region-level and pixel-level losses. As illustrated in the two line charts on the right of Figure~\ref{fig:lamda_ab}, using only region-level loss (\textit{i.e.}, 1:0) or only pixel-level loss (\textit{i.e.}, 0:1) yields inferior performance compared to a balanced combination of both. Based on these findings, we adopt $\lambda_{\text{region}} = 1\mathrm{e}{-3}$ and $\lambda_{\text{pixel}} = 5\mathrm{e}{-5}$ as the default configuration in all subsequent experiments.

\subsection{Mapping Table Setup} \label{appen3}
To align the text descriptions extracted by RAM with the 150 semantic categories defined by Mask2Former and enable RAM-generated descriptions to inherit Mask2Former’s segmentation annotations, we construct a mapping table. Specifically, we first apply RAM to extract labels across the entire training dataset. We then employ a sentence similarity model Sentence-BERT \cite{reimers-2019-sentence-bert} to compute the most semantically similar Mask2Former label for each RAM label and record the corresponding similarity score. Alignments with low confidence (similarity $\leq 0.5$) are discarded and additional unreasonable matches are manually filtered out. A partial mapping table is presented in Table~\ref{tab:word_sim}, and the complete version will be made publicly available in future releases.

\begin{table}[htbp]
\centering
\small
\caption{Partial mapping table display.}
\label{tab:word_sim}
\begin{tabular}{ccc}
\toprule
RAM-generated Word & Most Similar Word & Similarity Score↑ \\
\midrule
lake        & lake         & 1.0000 \\
blanket     & blanket      & 0.8754 \\
woman       & person       & 0.6375 \\
sit         & seat         & 0.7008 \\
stair       & stairs       & 0.9246 \\
home        & house        & 0.8043 \\
entrance    & door         & 0.6537 \\
doorway     & door         & 0.7985 \\
harbor      & pier         & 0.6539 \\
pond        & lake         & 0.7054 \\
... & ...    & ... \\
\bottomrule
\end{tabular}
\end{table}
% \begin{figure}[htbp]
% 	\centering
% 	\includegraphics[width=\linewidth]{AnonymousSubmission/LaTeX/appendix_mapping.jpg}
% 	\caption{Partial mapping table display.}
% 	\label{fig:mapping}
%     % \vspace{-0.3cm}
% \end{figure}

\subsection{Cross-Attention Map Visualization Analysis.} \label{appen4}
\begin{figure}[htbp]
	\centering
	\includegraphics[width=0.8\linewidth]{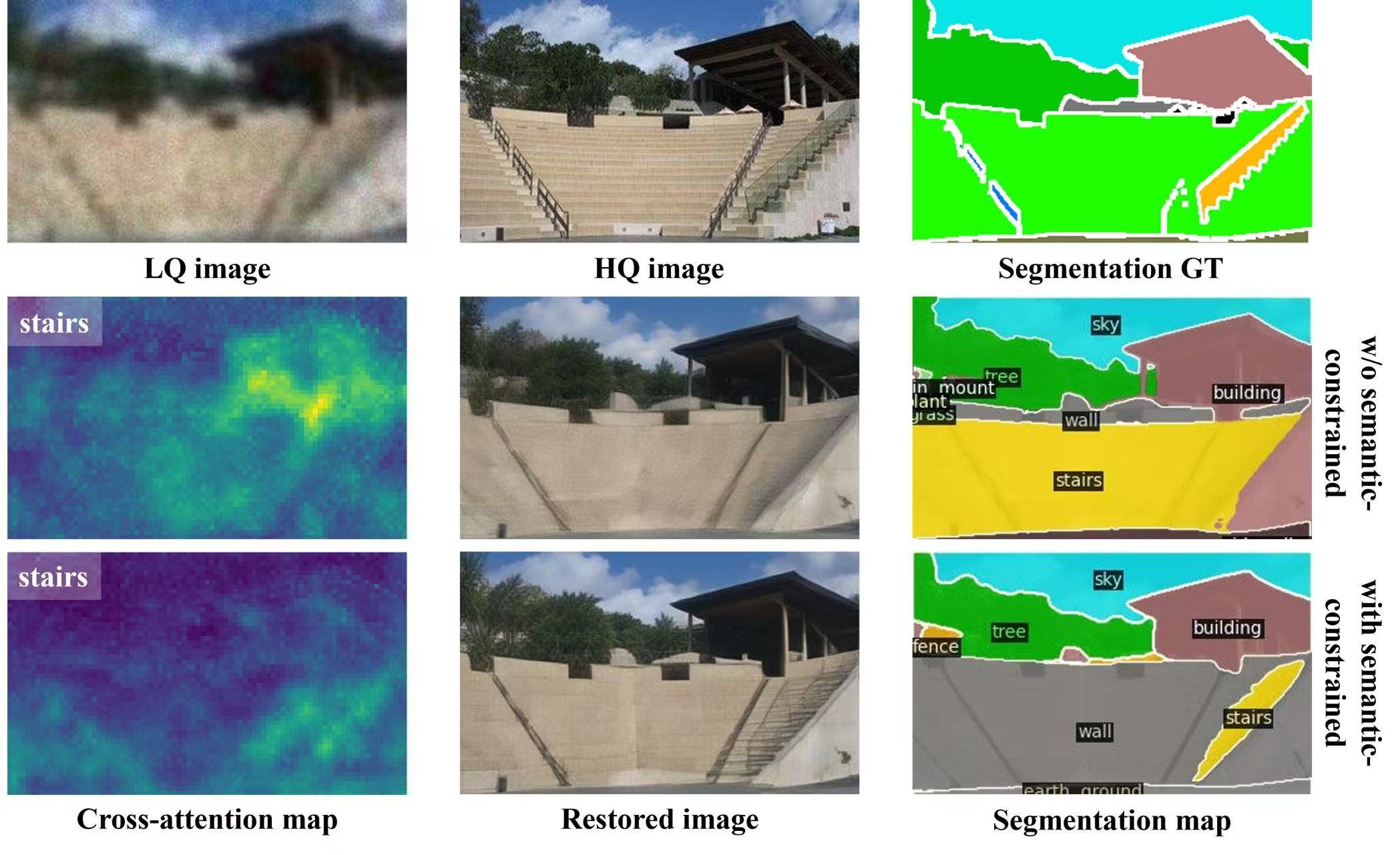}
	\caption{Cross-attention map visualization results. Sample is from ADE20K.}
	\label{fig:attn}
    % \vspace{-0.3cm}
\end{figure}

We visualize the cross-attention maps during the restoration process in Figure~\ref{fig:attn}. The first column shows the aggregated cross-attention maps, obtained by merging all attention layers in the latent diffusion network. The second column presents the restored images, and the third column shows the RASS segmentation results of the restored image. It can be clearly observed that the restoration model with semantic constraints exhibits more accurate attention to semantically meaningful regions, such as the `stairs' area (third row). In contrast, the model without semantic constraints shows inaccurate attention, \textit{e.g.}, responding to `building' as `stairs' area (second row), leading to coarse segmentation results. These observations highlight the effectiveness of SCL in enhancing the semantic fidelity of restoration and supporting downstream segmentation performance.

\subsection{Comparison Results} \label{appen5}
We present quantitative comparison results between SCR and GAN-based models. In Table \ref{tab:gan}, we compare SCR with four representative GAN-based Real-ISR methods: BSRGAN \cite{zhang2021designing}, Real-ESRGAN \cite{wang2021real}, LDL \cite{liang2022details}, and FeMaSR \cite{chen2022real}. Unsurprisingly, these GAN-based methods outperform SCR in terms of fidelity metrics such as PSNR and SSIM, while SCR achieves a clear advantage in perceptually limited metrics.

\begin{table*}[htbp]
\caption{Quantitative comparison with GAN-based methods on DrealSR, RealSR and DIV2K datasets. Best results are highlighted in \textbf{bold}.}

\label{tab:gan}
\centering
\small
\renewcommand{\arraystretch}{1.05}
\resizebox{\textwidth}{!}{
\begin{tabular}{l|l|cccccccc}
\toprule
Dataset & Method & PSNR↑ & SSIM↑ & DISTS↓ & NIQE↓ & MUSIQ↑ & MANIQA↑ & CLIPIQA↑ \\
\midrule
\multirow{5}{*}[0pt]{\centering DrealSR}
& BSRGAN \cite{zhang2021designing} & \textbf{26.39} & \textbf{0.7654} & 0.2121 & 5.6567 & 63.21 & 0.5399 & 0.5001 \\
& Real-ESRGAN \cite{wang2021real} & 25.69 & 0.7616 & \textbf{0.2063} & 5.8295 & 60.18 & 0.5487 & 0.4449 \\
& LDL \cite{liang2022details} & 25.28 & 0.7567 & 0.2121 & 6.0024 & 60.82 & 0.5485 & 0.4477 \\
& FeMASR \cite{chen2022real} & 25.07 & 0.7358 & 0.2288 & 5.7885 & 58.95 & 0.4865 & 0.5270 \\
& SCR (Ours) & 25.34 & 0.7219 & 0.2149 & \textbf{5.2387} & \textbf{70.27} & \textbf{0.6689} & \textbf{0.6718} \\
\midrule
\multirow{5}{*}[0pt]{\centering RealSR}
& BSRGAN \cite{zhang2021designing} & \textbf{26.39} & \textbf{0.7654} & 0.2121 & 5.6567 & 63.21 & 0.5399 & 0.5001 \\
& Real-ESRGAN \cite{wang2021real} & 25.69 & 0.7616 & \textbf{0.2063} & 5.8295 & 60.18 & 0.5487 & 0.4449 \\
& LDL \cite{liang2022details} & 25.28 & 0.7567 & 0.2121 & 6.0024 & 60.82 & 0.5485 & 0.4477 \\
& FeMASR \cite{chen2022real} & 25.07 & 0.7358 & 0.2288 & 5.7885 & 58.95 & 0.4865 & 0.5270 \\
& SCR (Ours) & 25.34 & 0.7219 & 0.2149 & \textbf{5.2387} & \textbf{70.27} & \textbf{0.6689} & \textbf{0.6718} \\
\midrule
\multirow{5}{*}[0pt]{\centering DIV2K}
& BSRGAN \cite{zhang2021designing} & \textbf{24.58} & 0.6269 & 0.2275 & 4.7518 & 61.20 & 0.5071 & 0.5247 \\
& Real-ESRGAN \cite{wang2021real} & 24.29 & \textbf{0.6371} & 0.2141 & 4.6786 & 61.06 & 0.5501 & 0.5277 \\
& LDL \cite{liang2022details} & 23.83 & 0.6344 & 0.2227 & 4.8554 & 60.04 & 0.5350 & 0.5180 \\
& FeMASR \cite{chen2022real} & 23.06 & 0.5887 & \textbf{0.2057} & 4.7410 & 60.83 & 0.5074 & 0.5997 \\
& SCR (Ours) & 23.73 & 0.6029 & 0.2231 & \textbf{4.5793} & \textbf{70.02} & \textbf{0.6561} & \textbf{0.6971} \\
\bottomrule
\end{tabular}
}
% \vspace{-0.2cm}
\end{table*}

We present quantitative comparison results between RASS and feature-refinement method in Table \ref{tab:ade20k_jafar_rass}, exemplified by JAFAR \cite{couairon2025jafar}, which improves DINOv2-ViT-S/14 \cite{oquab2023dinov2} (with bilinear upsampling) on SQ ADE20K from 39.23 to 40.49 mIoU. We consider three evaluation settings according to the quality of input images: Standard Quality (SQ), Simulated Degradation (SiD), and Real-World Degradation (RD). SQ images are from ADE20K; SiD images are obtained by applying the degradation pipeline of RealESRGAN to ADE20K; RD images are from our collected RealLQ dataset. We can see that RASS consistently outperforms the representative refinement based method JAFAR across SQ, SiD and RD settings, demonstrating that RASS is not limited to ``rescue'' cases but also yields large gains on standard quality and mild–moderate degradations. However, we agree that the cost of RASS is higher than JAFAR (FPS 1.84 vs 8.13). Overall, the experiments support RASS as a general-purpose recovery module with certain cost in latency.

%Although refinement-based methods enjoy a significant inference speed advantage (e.g., 8.13 FPS for JAFAR vs. 1.84 FPS for RASS on an NVIDIA RTX A6000, with a batch size of 1), the large accuracy gap under realistic corruptions highlights the necessity of incorporating a strong semantic generative prior.}
\begin{table*}[htbp]
\centering
\small
\renewcommand{\arraystretch}{1.1}
\caption{Quantitative comparison with representative feature-refinement baseline (JAFAR). All FPS measured on NVIDIA A6000.}
\label{tab:ade20k_jafar_rass}
\begin{tabular}{l|c|ccc|ccc|c}
\toprule
\multirow{2}{*}{Method}& ADE20K (SQ) & \multicolumn{3}{c|}{ADE20K (SiD)} & \multicolumn{3}{c|}{RealLQ (RD)} & \multirow{2}{*}{FPS} \\
\cmidrule(lr){2-2} \cmidrule(lr){3-5} \cmidrule(l){6-8}
 & DT & DT & R2S & FT & DT & R2S & FT  & \\
\midrule
%DINOv2-ViT-S/14 & 39.23 & 23.76 & 34.67 & 26.88 & 15.60 & 17.02 & 15.66 & 113.15 \\
JAFAR \cite{couairon2025jafar} & 40.49 & 23.91 & 34.82 & 28.01 & 17.77 & 18.40 & 20.23 & \textbf{8.13} \\
RASS (Ours) & \textbf{53.76} & - & \textbf{46.81} & \textbf{47.42} & - & \textbf{37.84} & \textbf{39.80} & 1.84 \\
\bottomrule
\end{tabular}
\end{table*}

\end{document}